%% file: iclr2026_conference.tex
\documentclass{article} 
\usepackage{iclr2026_conference,times}
\usepackage{multirow} 
\input{math_commands.tex}

\usepackage{booktabs} 
\usepackage{hyperref}
\usepackage{url}
\usepackage{graphicx} 
\usepackage{amssymb} 
\usepackage[ruled,vlined]{algorithm2e}
\usepackage{graphicx}
\usepackage{subcaption}
\usepackage{tikz}
\usepackage{pgf-pie}
\usepackage{caption} 
\usepackage{wrapfig}
\usepackage{xcolor}
\definecolor{DraftYellow}{RGB}{248,209,78}   
\definecolor{VerifBlue}{RGB}{47,120,216}     
\definecolor{PathGreen}{RGB}{45,190,97}      
\usepackage{pifont}
\usepackage[most]{tcolorbox}

\usepackage[most]{tcolorbox} 
\tcbset{
  colback=white, colframe=black!15, boxrule=0.4pt,
  arc=1.5mm, left=6pt, right=6pt, top=6pt, bottom=6pt
}
\newtcolorbox{vizbox}[1][]{breakable, enhanced, title=#1}

\usepackage{amssymb}   
\title{DiffuSpec: Unlocking Diffusion Language Models for Speculative Decoding}

\author{
  Guanghao Li\textsuperscript{1,2} \quad
  Zhihui Fu\textsuperscript{3} \quad
  Min Fang\textsuperscript{3} \quad
  Qibin Zhao\textsuperscript{3} \quad
  \textbf{Ming Tang}\textsuperscript{2,\thanks{Corresponding author}} \and
  \textbf{Chun Yuan}\textsuperscript{1,\footnotemark[1]} \quad
  \textbf{Jun Wang}\textsuperscript{3} \\[6pt]
  \textsuperscript{1}SIGS, Tsinghua University \quad
  \textsuperscript{2}Southern University of Science and Technology \\
  \textsuperscript{3}OPPO Research Institute
}

\iclrfinalcopy 
\begin{document}

\maketitle
\def\thefootnote{}
\footnotetext{Email: ligh24@mails.tsinghua.edu.cn}
\def\thefootnote{\arabic{footnote}}

\input{sec/0_abstract}

\input{sec/1_intro}

\input{sec/5.relatedwork}

\input{sec/2.Preliminaries}

\input{sec/3.method}

\input{sec/4.experiments}

\input{sec/6.conclusion}


\input{sec/appendix}

\end{document}

%% file: math_commands.tex
\usepackage{amsmath,amsfonts,bm}

\def\eqref#1{equation~\ref{#1}}

\def\1{\bm{1}}

\DeclareMathAlphabet{\mathsfit}{\encodingdefault}{\sfdefault}{m}{sl}
\SetMathAlphabet{\mathsfit}{bold}{\encodingdefault}{\sfdefault}{bx}{n}



%% file: sec/0_abstract.tex
\begin{abstract}
As large language models (LLMs) scale up, accuracy improves, but the autoregressive (AR) nature of decoding increases latency since each token requires a serial forward pass. 
Speculative decoding addresses this by employing a fast drafter to propose multi-token drafts, which are then verified in parallel by the target model.
However, many deployments still rely on AR drafters, where sequential passes limit wall-clock gains. We revisit the drafting stage and present \textbf{DiffuSpec}, a training-free drop-in framework that uses a pretrained diffusion language model (DLM) to produce multi-token drafts in a single forward pass, while remaining compatible with standard AR verifiers. 
Because DLM drafts are generated under bidirectional conditioning, parallel per-position candidates form a token lattice in which the locally highest-probability token at each position need not form a causal left-to-right path. 
Moreover, DLM drafting requires pre-specifying a draft length, inducing a speed-quality trade-off.  
To address these challenges, we introduce two practical components: (i) a causal-consistency path search (CPS) over this lattice that extracts a left-to-right path aligned with AR verification; and (ii) an adaptive draft-length (ADL) controller that adjusts next proposal size based on recent acceptance feedback and realized generated length. Across benchmarks, DiffuSpec yields up to $3\times$ wall-clock speedup, establishing diffusion-based drafting as a robust alternative to autoregressive drafters for speculative decoding.
\end{abstract}

%% file: sec/1_intro.tex
\section{Introduction}
\label{sec:intro}

Large language models (LLMs) continue to improve with scale, yet autoregressive (AR) decoding remains a latency bottleneck because generating $K$ tokens requires $K$ serial forward passes \citep{leviathan2023fast,hoffmann2022training}. A common line of work accelerates inference via pruning and sparsity, quantization, or knowledge distillation, but these techniques often introduce accuracy trade-offs or additional engineering complexity \citep{frantar2022gptq,frantar2023sparsegpt,xu2024survey}. Speculative decoding offers a nearly lossless alternative: a fast drafter first proposes multi-token drafts, and then the target model verifies the drafts in parallel, which preserves the target distribution while reducing wall-clock time \citep{xia2024unlocking}. However, the speedup hinges on two factors: the drafter’s per-step drafting throughput and the verification acceptance rate, defined as the fraction of drafted tokens accepted by the AR verifier during parallel verification.

\begin{figure}[t]
  \centering
  \includegraphics[width=\linewidth]{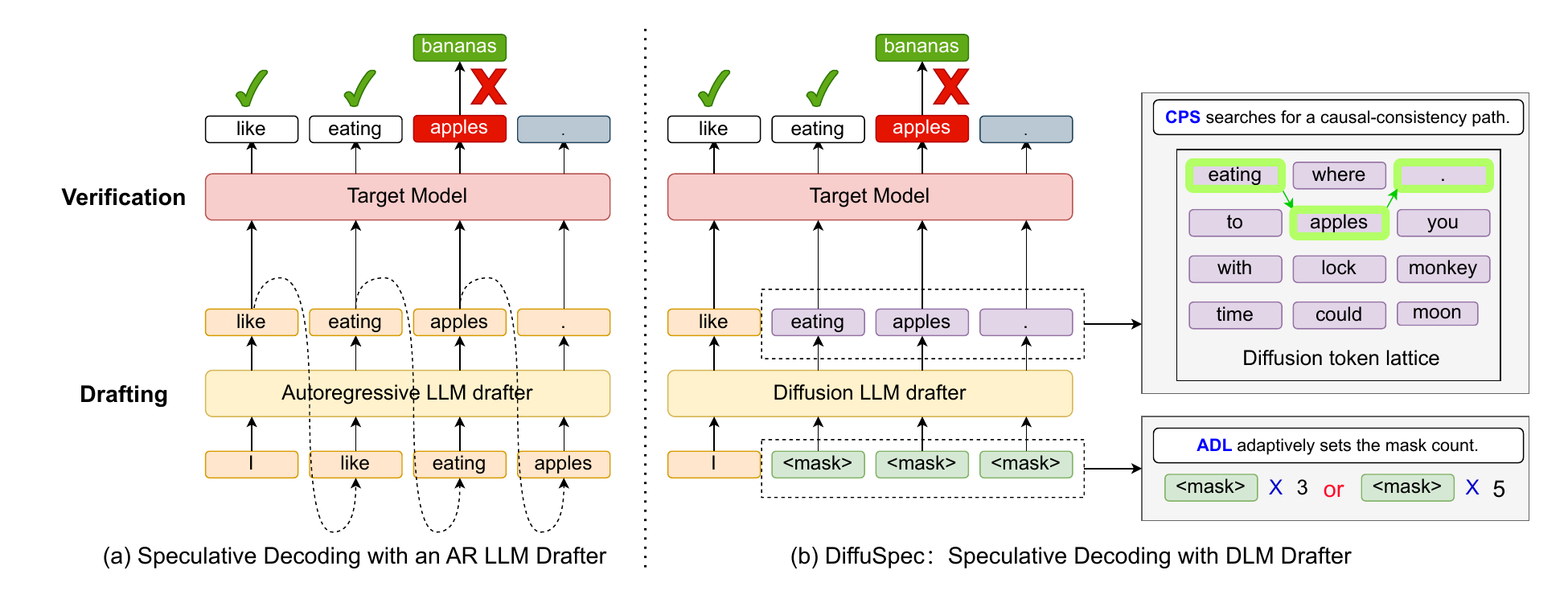}
  \vspace{-0.9cm}
  \caption{\textbf{Speculative decoding: AR vs.\ DiffuSpec.}
  (a) \textbf{AR drafter:} drafts are produced sequentially and then block-verified by the target AR model.
  (b) \textbf{DiffuSpec (DLM drafter):} a single forward pass proposes a block for one-shot parallel verification; within DiffuSpec, causal-consistency path search (CPS) selects a left-to-right path from the diffusion token lattice, and the adaptive draft-length (ADL) controller sets the next draft length by selecting how many masked positions to fill.}

  \label{fig:diffuspec-framework}
  \vspace{-0.3cm}
\end{figure}

In practice, most deployments still use a small AR drafter (Fig.~\ref{fig:diffuspec-framework}a), which remains sequential and therefore pays one forward pass per drafted token, diluting the gains of parallel verification \citep{leviathan2023fast,chen2023accelerating}. Block prediction variants attach auxiliary heads to forecast future tokens in chunks, but they require extra training and effectively cap the maximum accepted length by the head depth or branching design, limiting end-to-end acceleration \citep{cai2024medusa}. Recent EAGLE-style methods rethink the drafter–target interface and achieve strong improvements with lightweight training or calibration, yet they still introduce additional learned components and deployment-time tuning \citep{li2024eagle,li2024eagle2,li2025eagle3}. 

Recent advances in diffusion language models (DLMs) \citep{li2022diffusion,austin2021structured} open a new avenue for speculative decoding. Several pre-trained DLMs (Fig.~\ref{fig:diffuspec-framework}b) can propose a block of token candidates in a single forward pass and optionally refine them iteratively \citep{nie2025large,ye2025dream}. These capabilities directly match drafter desiderata—higher per-step drafting throughput and strong proposal quality—making DLMs a compelling fit for parallel generation with parallel verification. However, DLM proposals are generated under bidirectional conditioning rather than strict left-to-right causality. This induces a  diffusion token lattice whose nodes are per-position candidates, where locally highest-probability token need not define a causal left-to-right path. In addition, DLM drafting requires a preset draft length. Together, these properties raise two practical questions we study: (i) \textbf{causal alignment}: how to select, from this lattice, a left-to-right path aligned with AR verification to maximize acceptance; and (ii) \textbf{draft length}: how to choose the block size to balance drafting cost against verification acceptance, since longer drafts increase proposal cost without guaranteeing higher acceptance.
 While concurrent work has begun to explore diffusion-based drafters \citep{christopher2024speculative}, a training-free drop-in framework with a systematic treatment of causal consistency and draft length remains under-explored.

To address these two issues, we present \textbf{DiffuSpec}, a training-free drop-in speculative decoding framework that uses a pretrained DLM as the drafter. DiffuSpec has two components: (i) a \emph{causal-consistency path search} (CPS) over the diffusion token lattice that selects a left-to-right path aligned with AR verification to maximize acceptance; and (ii) an \emph{adaptive draft-length} (ADL) controller that sets the next draft length using recent acceptance and realized generation length. DiffuSpec requires no additional training or architectural changes to the target model and integrates as a drop-in drafter via existing interfaces, with minimal serving-stack adjustments. Across diverse generation tasks, DiffuSpec delivers up to $3\times$ wall-clock speedup, outperforming other training-free baselines and approaching training-based methods. 

In summary, our main contributions include:

\begin{itemize}
  \item We introduce pretrained DLMs as drafters for speculative decoding and analyze two defining traits—bidirectional conditioning and preset draft length—showing how they jointly affect verifier acceptance and end-to-end speedup and what challenges they pose.

  \item We introduce \textbf{DiffuSpec}, a training-free drop-in drafter that (i) performs CPS to align proposals with AR verification and boost acceptance, and (ii) uses an ADL controller to choose the next draft length near the speed--quality sweet spot; DiffuSpec integrates with existing AR verifiers with minimal serving-stack adjustments.
  \item We demonstrate that DiffuSpec achieves up to $3\times$ wall-clock speedup across tasks, surpassing training-free baselines and approaching training-based methods, thereby establishing the viability of DLMs as effective drafters for speculative decoding.
\end{itemize}

%% file: sec/5.relatedwork.tex
\section{Related Work}
\label{sec:relatedwork}

\textbf{Speculative decoding.}
Speculative decoding accelerates autoregressive (AR) generation by letting a fast \emph{drafter} propose multiple tokens that a target LM verifies in parallel, while preserving the target distribution~\citep{xia2024unlocking,sun2025scaling}. 
\emph{Training-free} variants either use a smaller pretrained AR drafter~\citep{leviathan2023fast,chen2023accelerating} or \emph{retrieval/cache}-based drafters that mine recent $n$-grams or suffix structures~\citep{he2023rest,saxena2023prompt}, and are complemented by verification-side improvements such as block verification and massively parallel cache-tree validation~\citep{sun2024block,miao2024specinfer,svirschevski2024specexec}. 
A separate line reduces strict step-by-step dependency without an auxiliary drafter via \emph{lookahead} updates~\citep{fu2024break}. 
\emph{Training-based} variants either attach multi-token prediction (MTP) heads to the target LM~\citep{cai2024medusa,ankner2024hydra} or distill a separate drafter that operates at the feature/token level~\citep{li2024eagle,li2024eagle2,li2025eagle3}. Training-based methods can attain high acceptance, but they incur additional training and maintenance; retrieval-based drafters may be domain-sensitive and can fail on short matches. Our goal is a \emph{training-free} drafter with high per-step throughput and robust acceptance.

\textbf{Diffusion language models.}
Discrete/latent diffusion for text spans early D3PMs~\citep{austin2021structured} and Diffusion-LM~\citep{li2022diffusion} to hybrids with PLMs~\citep{zhou2023diffusion,he2022diffusionbert} and recent scaling/adaptation frameworks~\citep{gong2024scaling}. 
From-scratch large DLMs report competitiveness with similarly sized AR baselines while retaining diffusion-style parallel refinement~\citep{nie2025large,ye2025dream}. 
At inference time, DLMs natively support parallel multi-token updates with iterative refinement but pay for bidirectional attention and multiple denoising steps; accordingly, training-free accelerators (adaptive KV caching, dynamic cache eviction, suffix-dropout pruning) have emerged~\citep{liu2025dllm,song2025sparse,chen2025dpad}. 
These traits—single-pass proposal of token blocks  and strong proposal quality—make DLMs promising drafters for speculative decoding.

\textbf{Diffusion as a drafter for speculative decoding.}
\citet{christopher2024speculative} first showed that a discrete diffusion model can draft sequences for AR verification, validating the feasibility of diffusion-based drafting. 
However, prior work typically (i) trains or calibrates a dedicated diffusion drafter and (ii) lacks a systematic analysis of how draft length and the diffusion-induced token lattice with relaxed causality interact with AR verification. 
In contrast, \textbf{DiffuSpec} is \emph{training-free} and introduces (a) a \emph{causal-consistency path search} (CPS) over the diffusion-induced token lattice and (b) an \emph{adaptive draft-length} (ADL) controller to maximize accepted prefixes under AR block verification, yielding strong wall-clock speedups.

%% file: sec/2.Preliminaries.tex
\section{Preliminaries—Speculative Decoding}
\label{sec:prelim}

Let $p_\theta$ be the target autoregressive (AR) language model and $\mathbf{x}_{1:j}$ the current prefix.
Speculative decoding~\citep{leviathan2023fast,chen2023accelerating,xia2024unlocking} accelerates generation under a \emph{drafter–verifier} interface: a fast drafter proposes a short continuation, and the target AR model verifies it in parallel while preserving the $p_\theta$ distribution.

\textbf{Drafting.}
Given $\mathbf{x}_{1:j}$, a drafter $q_\phi$ proposes a length-$k_t$ block
$\hat{\mathbf{y}}_{j+1:j+k_t}=(\hat{y}_{j+1},\ldots,\hat{y}_{j+k_t})$ conditioned on $\mathbf{x}_{1:j}$,
and records per-position conditional probabilities
$\{q_\phi(\hat{y}_{j+i}\mid \mathbf{x}_{1:j+i-1})\}_{i=1}^{k_t}$. Here $t=1,2,\ldots$ indexes speculative steps.

\textbf{Parallel verification.}
The target model evaluates the drafted tokens in a single parallel pass, producing
$\{p_\theta(\hat y_{j+i}\mid \mathbf{x}_{1:j+i-1})\}_{i=1}^{k_t}$,
and then processes them left-to-right with the standard acceptance rule:
\begin{equation}
\alpha_{t,i}\;=\;\min\!\left(1,\;
\frac{p_\theta(\hat y_{j+i}\mid \mathbf{x}_{1:j+i-1})}
     {q_\phi(\hat y_{j+i}\mid \mathbf{x}_{1:j+i-1})}\right),
\quad i=1,\ldots,k_t.
\end{equation}
If $\hat y_{j+i}$ is rejected, a replacement is sampled from the residual distribution proportional to 
$\big[p_\theta(\cdot\mid \mathbf{x}_{1:j+i-1})-q_\phi(\cdot\mid \mathbf{x}_{1:j+i-1})\big]_+$, 
where $[u]_+=\max(u,0)$, followed by normalization; all remaining drafted tokens are discarded before continuing.
This procedure is unbiased with respect to $p_\theta$~\citep[App.~A.1]{leviathan2023fast} and admits verifier-side engineering such as block or tree-based parallel verification to further reduce latency~\citep{sun2024block,miao2024specinfer}.

\textbf{Accepted prefix length.}
At speculative step $t$ with proposal length $k_t$, let $A_{t,i}\!\in\!\{0,1\}$ indicate whether the $i$-th drafted token is accepted by the verifier \emph{given} that positions $1{:}i\!-\!1$ were accepted.
The number of tokens actually committed is
\begin{equation}
L^{\mathrm{acc}}_t
\;=\;
\max\!\big\{m\in\{0,\ldots,k_t\}:\;A_{t,1}=\cdots=A_{t,m}=1\big\}
\;=\;
\sum_{i=1}^{k_t}\prod_{r=1}^{i} A_{t,r}.
\end{equation}
The verifier appends the accepted prefix $\hat{\mathbf{y}}_{j+1:j+L^{\mathrm{acc}}_t}$ and discards the remainder, yielding the updated prefix $\mathbf{x}_{1:j+L^{\mathrm{acc}}_t}$. Decoding terminates early if an $\mathrm{EOS}$ token is accepted. We use $L^{\mathrm{acc}}_t$ as a per-step measure of useful progress; holding latency fixed, larger values imply higher speedup.

%% file: sec/3.method.tex
\section{DiffuSpec}
\label{sec:method}

As shown in Fig.~\ref{fig:diffuspec-framework}b, \textbf{DiffuSpec} departs from conventional speculative decoding by replacing the AR drafter with a pretrained diffusion language model (DLM) that proposes a length-$k_t$ draft in a single forward pass, and by augmenting drafting with \emph{causal-consistency path search} (CPS) and an \emph{adaptive draft-length} (ADL) controller. We next describe these three components in turn.

\subsection{DLM As A Training-Free Drafter}
\label{sec:method:dlm}

Unlike autoregressive models with fixed left-to-right factorization, diffusion language models learn a non-autoregressive denoising mapping that reconstructs clean text from corrupted text~\citep{austin2021structured,gong2024scaling,nie2025large,ye2025dream,chen2025dpad}.

\textbf{Training.}
Let $\mathbf{x}^{(0)}$ be a clean sequence and $\mathbf{x}^{(\eta)}$ its corrupted counterpart at noise level $\eta\in[0,1]$.
We define a forward corruption kernel $r$ with a user-specified discrete noise prior $\pi_{\text{noise}}$:
\begin{equation}
r\!\left(x_i^{(\eta)} \mid x_i^{(0)}\right)=
(1-\eta)\,\mathbf{1}\{x_i^{(\eta)}=x_i^{(0)}\}
+\eta\,\pi_{\text{noise}}\!\left(x_i^{(\eta)}\right),
\end{equation}
where $\sum_v \pi_{\text{noise}}(v)=1$ (e.g., all mass on \texttt{[MASK]} or a mixture over noise symbols).
A parameterized denoiser $q_\phi$ is trained with token-wise cross-entropy to predict originals at corrupted positions:
\begin{equation}
\mathcal{L}(\phi)=
-\,\mathbb{E}_{\eta,\mathbf{x}^{(0)},\mathbf{x}^{(\eta)}}\!
\Bigg[\sum_{i:\,x_i^{(\eta)}\neq x_i^{(0)}}
\log q_\phi\!\big(x_i^{(0)} \mid \mathbf{x}^{(\eta)}\big)\Bigg],
\end{equation}
where $q_\phi$ is a Transformer with bidirectional attention.

\textbf{Inference (iterative refinement).}
Given a prefix $\mathbf{r}=\mathbf{x}_{1:j}$ and target length $k_t$, initialize
$\mathbf{y}^{(0)}=\mathbf{r}\circ(\texttt{[MASK]})^{k_t}$ with masked set $M_0=\{j{+}1,\ldots,j{+}k_t\}$, where $\circ$ denotes concatenation.
For refinement steps $s=1,\ldots,S$, compute per-position conditionals $q_\phi(y_i\mid \mathbf{y}^{(s-1)})$ for $i\in M_{s-1}$, choose an update subset $U_s\subseteq M_{s-1}$ (e.g., top-$K$ by confidence), and set
\begin{equation}
y_i^{(s)}=\begin{cases}
\arg\max_{v\in\mathcal{V}} q_\phi\!\big(y_i{=}v \mid \mathbf{y}^{(s-1)}\big), & i\in U_s,\\
y_i^{(s-1)}, & \text{otherwise},
\end{cases}
\qquad
M_s=M_{s-1}\setminus U_s,
\end{equation}
until $M_s=\varnothing$. By default we use a single refinement pass ($S{=}1$) to isolate drafting cost; $S{>}1$ is ablated in Sec.~\ref{sec:experiments}.

\textbf{Integration with speculative decoding.}
At speculative step $t$ with prefix $\mathbf{x}_{1:j}$, a pretrained DLM proposes a length-$k_t$ block
$\hat{\mathbf{y}}_{j+1:j+k_t}$ in essentially one orward/refinement  pass, and can expose per-position top-$M$ candidate sets with log-scores taken under the current draft context $\mathbf{y}^{(S)}$.
For verifier-side acceptance with a DLM drafter, we evaluate a left-to-right proxy by masking all future draft positions when scoring the token at $j{+}i$:
\begin{equation}
q_\phi^{\mathrm{L2R}}(v\mid \mathbf{x}_{1:j+i-1})
:=\;q_\phi\!\Big(v\;\Big|\;\mathbf{x}_{1:j}\circ \underbrace{(\texttt{[MASK]})^{i-1}}_{\text{past in-block}},\, 
\underbrace{(\texttt{[MASK]})^{k_t-i+1}}_{\text{future in-block}}\Big).
\end{equation}
We use $q_\phi^{\mathrm{L2R}}$ in the standard acceptance ratio. Accordingly, Sec.~\ref{sec:method:cps} introduces \textbf{CPS} to align proposals causally with the AR verifier, and Sec.~\ref{sec:method:adl} presents \textbf{ADL} to set $k_t$ near the speed–quality sweet spot.

\begin{figure}[t]
  \centering
  \begin{minipage}{0.48\linewidth}
    \centering
    \includegraphics[width=\linewidth]{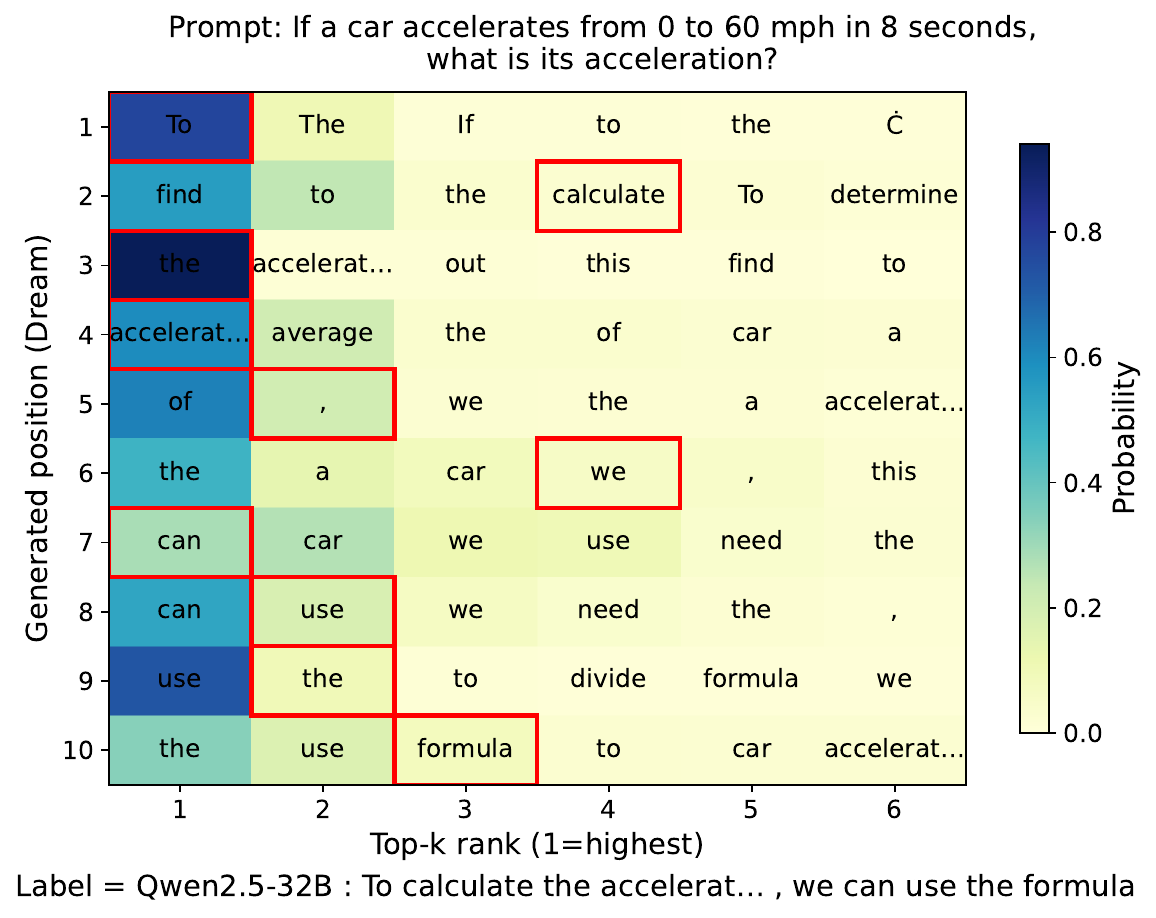}
    \captionof{figure}{\textbf{DLM token-mass diffusion (Dream-7B).}
      Probability mass spreads across positions during joint block refinement; the per-position top-1 need not yield an AR-consistent left-to-right path under $p_\theta$.}
    \label{fig:topk-heatmap}
  \end{minipage}\hfill
  \begin{minipage}{0.48\linewidth}
    \centering
    \includegraphics[width=\linewidth]{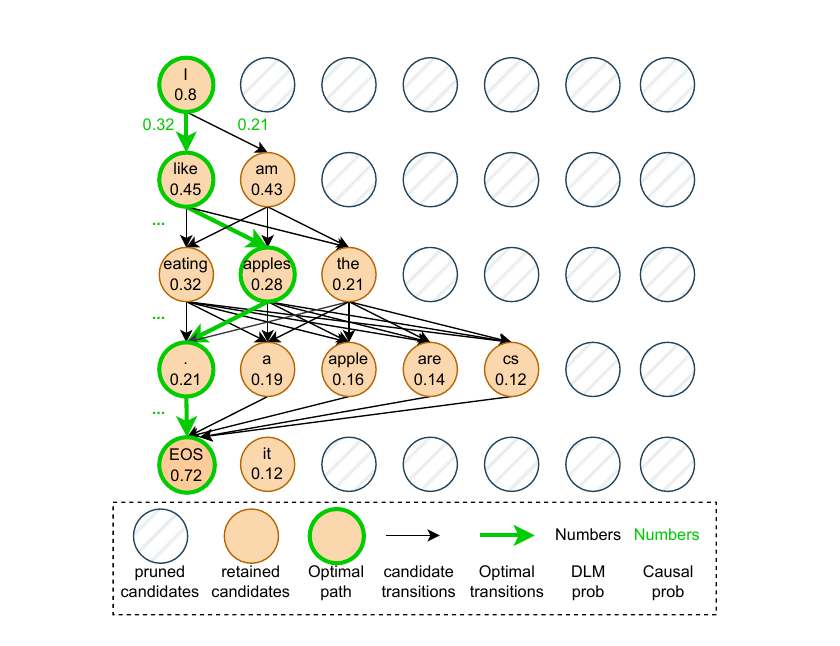}
    \captionof{figure}{\textbf{Pruned candidate lattice and CPS.}
      We keep tokens via a cumulative-mass threshold $\tau$ (e.g., $0.8$), always retain $\mathrm{EOS}$, early-stop after the first $\mathrm{EOS}$, and select the best path using a DLM score plus a causal ($n$-gram) proxy.}
    \label{fig:pruned-search}
  \end{minipage}
  \vspace{-0.4em}
\end{figure}

\subsection{Causal-Consistency Path Search (CPS)}
\label{sec:method:cps}

\textbf{Phenomenon and motivation.}
Under relaxed causality, the DLM refines tokens jointly within a block. As a result, token probability mass spreads across positions and the per-position top-1 chosen by the DLM is not necessarily the best left-to-right choice for the AR verifier $p_\theta$ (Fig.~\ref{fig:topk-heatmap}). To mitigate this mismatch, we explicitly search before verification—for a left-to-right path that is both high-confidence under the DLM and fluent under a causal proxy (Fig.~\ref{fig:pruned-search}).

\begin{figure}[t]
  \centering
  \begin{minipage}{0.48\linewidth}
    \centering
    \includegraphics[width=\linewidth]{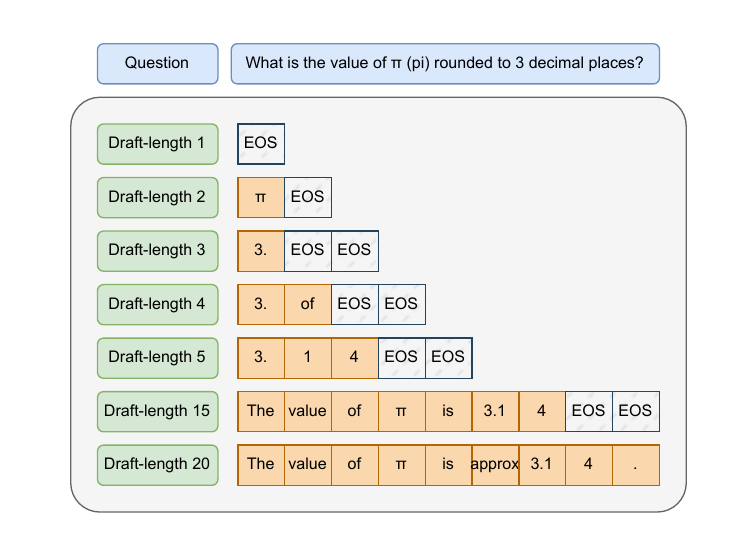}
    \captionof{figure}{\textbf{Qualitative effect of draft length.}
      As the draft length $k_t$ increases, DLM proposals evolve from short fragments to more complete answers; once the model deems the content ``complete,'' an early \textsc{eos} truncates further content.}
    \label{fig:causal-schematic}
  \end{minipage}\hfill
  \begin{minipage}{0.48\linewidth}
    \centering
    \includegraphics[width=\linewidth]{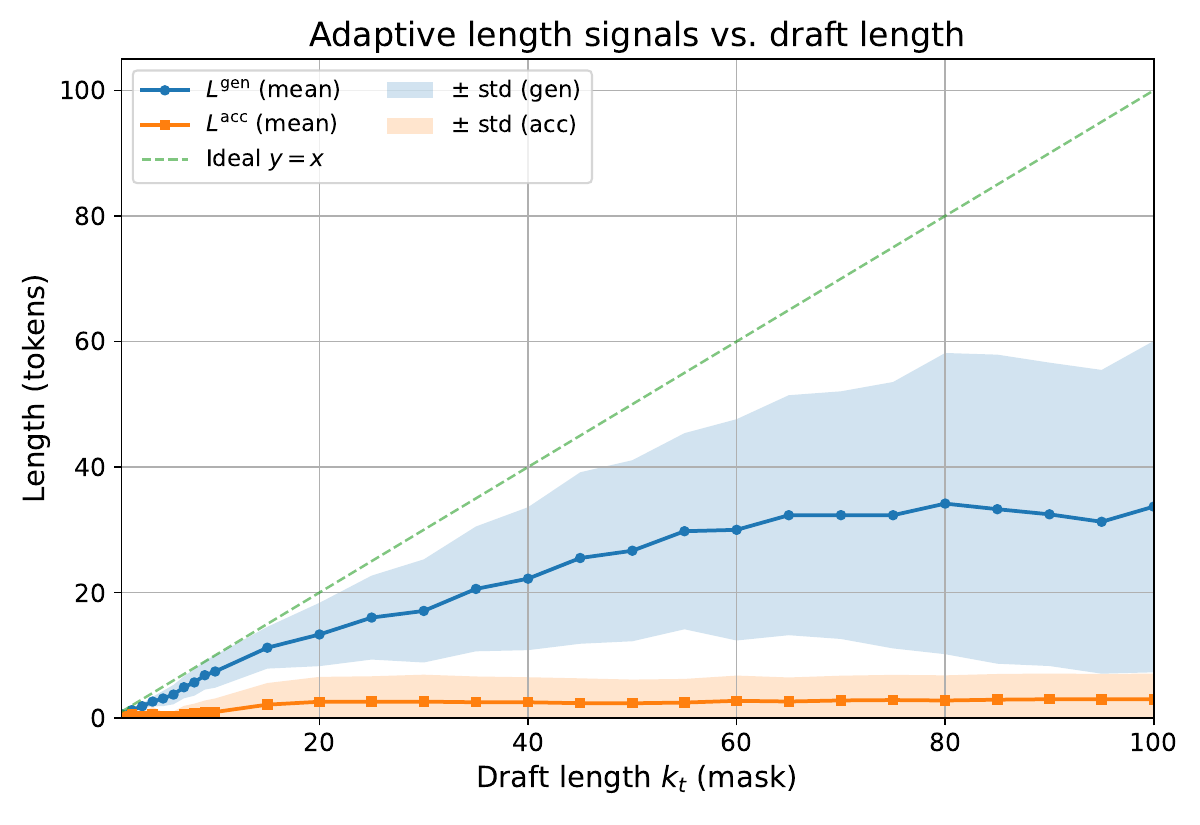}
    \captionof{figure}{\textbf{Adaptive-length signals vs.\ draft length.}
      For each $k_t$, we plot the mean and $\pm$1 standard deviation  of the $\mathrm{EOS}$-aware generation length $L^{\mathrm{gen}}$ and the accepted length $L^{\mathrm{acc}}$ across evaluation prompts. The dashed diagonal $y{=}x$ marks the ideal should-generate line.}
    \label{fig:adaptive-length}
  \end{minipage}
  \vspace{-0.4em}
\end{figure}

\textbf{Lattice and pruning.}
We first specify the search space. From the final DLM pass, for each position $i=1{:}k_t$ we extract a candidate set $\mathcal{C}_{j+i}$ (top-$M$) with log-scores
$\ell^{\mathrm{dlm}}_{j+i}(v)=\log q_\phi\!\big(v\mid \mathbf{x}_{1:j},\mathbf{y}^{(S)}_{\setminus (j+i)}\big)$,
i.e., conditioning on the current draft context except the target position. The naive Cartesian product over $\{\mathcal{C}_{j+i}\}_{i=1}^{k_t}$ is exponential, so we apply a training-free, mass-adaptive pruning rule that respects local uncertainty. 
Let $p_{j+i}(v)=\exp(\ell^{\mathrm{dlm}}_{j+i}(v))$.
We retain the smallest prefix exceeding a cumulative-mass threshold $\tau$:
\begin{equation}
M_i=\min\Big\{m\le M_{\max}:\ \sum_{v\in \text{Top-}m} p_{j+i}(v)\ge\tau\Big\},\qquad
\mathcal{C}_{j+i}\leftarrow\text{Top-}M_i.
\end{equation}
This makes $|\mathcal{C}_{j+i}|$ entropy-adaptive—peaky positions keep few candidates; flatter ones keep more, capped by $M_{\max}$. In addition, we stop expanding once the first $\mathrm{EOS}$ is placed: diffusion proposals tend to pad with $\mathrm{EOS}$ after the content is “complete” (qualitative trend in Fig.~\ref{fig:causal-schematic}), so exploring beyond the first $\mathrm{EOS}$ rarely yields causal gains.

\textbf{Scoring and search.}
Let $m_{\max}$ denote the depth up to (and including) the first $\mathrm{EOS}$ encountered during expansion. Given the pruned lattice, we score $\pi=(\pi_1,\ldots,\pi_m)$ by combining DLM confidence with a small causal proxy (e.g., an $n$-gram or a tiny causal LM):
\begin{equation}
\label{eq:cps:score}
\mathcal{S}(\pi)=\sum_{i=1}^m\Big[\lambda\,\ell^{\mathrm{dlm}}_{j+i}(\pi_i)\;+\;(1-\lambda)\,\ell^{\mathrm{ng}}_{j+i}(\pi_{1:i})\Big],
\end{equation}
where $\ell^{\mathrm{ng}}_{j+i}$ is the causal proxy log-score of $\mathbf{x}_{1:j}\!\circ\!\pi_{1:i}$ and $\lambda\in[0,1]$ trades off between DLM confidence and causal fluency.
We then run left-to-right beam search (beam $B$) on the pruned lattice until $\mathrm{EOS}$ is placed. If $\bar C$ denotes the average branching factor after pruning, the per-step complexity is $O(B\,\bar C\,m_{\max})$. As $\tau\!\to\!1$ and $B$ increases, the result approaches the unpruned optimum.result approaches the unpruned optimum.

\textbf{Effect.}
By entropy-adaptive pruning, early stopping at the first $\mathrm{EOS}$, and the causal–denoising score in \eqref{eq:cps:score}, CPS pushes the first $p_\theta$–$q_\phi$ mismatch farther to the right, thereby increasing the expected accepted length $L^{\mathrm{acc}}_t$ and improving end-to-end speed.

\subsection{Adaptive Draft Length (ADL)}
\label{sec:method:adl}

\textbf{Phenomenon and motivation.}
Draft length $k_t$ jointly determines drafting cost, proposal quality, and verifier acceptance.
Short drafts often yield terse fragments; moderate drafts capture more complete reasoning; very long drafts saturate content and trigger early EOS while also accumulating off-path tokens that the verifier rejects (Fig.~\ref{fig:causal-schematic}).
Empirically, the $\mathrm{EOS}$-aware generation length $L^{\mathrm{gen}}$ increases with $k_t$ and then saturates, and the accepted length $L^{\mathrm{acc}}$ tracks it (Fig.~\ref{fig:adaptive-length}).
The saturation point, however, is instance dependent and varies across prompts and along the trajectory, leading to large variance as shown in Fig.~\ref{fig:adaptive-length}.
A fixed $k_t$ therefore either wastes compute when too long or throttles progress when too short, which motivates an adaptive controller.

\textbf{Signals.}
Given the drafted block $\hat{\mathbf{y}}_{j+1:j+k_t}$, let $s_t$ be the index of the first $\mathrm{EOS}$ (or $+\infty$ if none) and define the $\mathrm{EOS}$-aware generation signal
\begin{equation}
L^{\mathrm{gen}}_t=\min(s_t-1,\,k_t).
\end{equation}
We compute $s_t$ from the raw DLM draft before CPS; since CPS also early-stops at the first $\mathrm{EOS}$, both signals are aligned.
After parallel verification we obtain the accepted prefix length $L^{\mathrm{acc}}_t$ as defined in Sec.~\ref{sec:prelim}.
To reduce volatility from occasional early \textsc{eos} or transient rejections, we use exponential moving averages:
\begin{equation}
\tilde L^{\mathrm{gen}}_t=(1-\rho)\tilde L^{\mathrm{gen}}_{t-1}+\rho L^{\mathrm{gen}}_t,\qquad
\tilde L^{\mathrm{acc}}_t=(1-\rho)\tilde L^{\mathrm{acc}}_{t-1}+\rho L^{\mathrm{acc}}_t,\qquad \rho\in(0,1].
\end{equation}

\textbf{Controller.}
With guardrails $k_{\min}\le k_{t+1}\le k_{\max}$, we adopt a one-line $O(1)$ policy:
\begin{equation}
\label{eq:adl}
k_{t+1} \;=\; \mathrm{clip}\!\Big(\,\big\lceil \tilde L^{\mathrm{gen}}_t + \delta\,\mathbf{1}\{\tilde L^{\mathrm{acc}}_t \ge \tilde L^{\mathrm{gen}}_t\}\big\rceil,\; 
k_{\min},\; k_{\max}\Big),
\end{equation}
where $\mathrm{clip}(z,a,b)=\min\{\max\{z,a\},b\}$ and $\delta>0$ is a small growth increment that activates when the verifier keeps up, namely when the accepted length matches the generated length on average.
Intuitively, $\tilde L^{\mathrm{gen}}_t$ estimates how much content the DLM is ready to produce before EOS, and $\tilde L^{\mathrm{acc}}_t$ indicates whether those tokens are reliably accepted; the policy increases $k_t$ only when both signals align.

\textbf{Effect.}
ADL tracks the instance-specific speed–quality sweet spot in real time.
As $k_t$ grows into the saturation regime, $L^{\mathrm{gen}}_t$ plateaus and the controller stabilizes; when acceptance lags, the policy avoids oversizing drafts; when acceptance catches up, it expands gently via $\delta$.

\begin{algorithm}[t]
\caption{DIFFUSPEC (4-stage): DLM drafting + CPS + parallel verification + ADL}
\label{alg:diffuspec}
\DontPrintSemicolon
\KwIn{prefix $\mathbf{x}_{1:j}$; target LM $p_\theta$; DLM $q_\phi$; ADL params $(k_{\min},k_{\max},\delta,\rho)$; CPS params $(M_{\max},\tau,B,\lambda)$.}
\textbf{Init:} $\tilde L^{\mathrm{gen}}_0\!\leftarrow\!0$, $\tilde L^{\mathrm{acc}}_0\!\leftarrow\!0$; set $k_1\!\leftarrow\!k_{\max}$.\;
\For{$t=1,2,\ldots$ until termination}{
  \textbf{(1) Draft:} run DLM to produce a length-$k_t$ block and per-position candidate sets $\{\mathcal{C}_{j+i}\}_{i=1}^{k_t}$ (top-$M_{\max}$ ) with scores $\ell^{\mathrm{dlm}}_{j+i}(\cdot)$.\;
  \textbf{(2) CPS:} on a pruned candidate lattice (cumulative-mass $\tau$, always keep $\mathrm{EOS}$, early-stop after the first $\mathrm{EOS}$), run left-to-right beam search (beam $B$) using score $\mathcal{S}(\cdot)$ in \eqref{eq:cps:score} to obtain a left-to-right path $\hat{\mathbf{y}}_{j+1:j+m_t}$ \textit{(path length $m_t$)}.\;
  \textbf{(3) Parallel verification:} block verification of $\hat{\mathbf{y}}_{j+1:j+m_t}$ with $p_\theta$; compute acceptance using $q_\phi^{\mathrm{L2R}}$; obtain $L^{\mathrm{acc}}_t$; append the accepted prefix and update $j\!\leftarrow\!j+L^{\mathrm{acc}}_t$; if an $\mathrm{EOS}$ is accepted, \textbf{terminate}.\;
  \textbf{(4) ADL:} compute $L^{\mathrm{gen}}_t$ from the proposal’s first-$\mathrm{EOS}$ index $s_t$; update EMAs $\tilde L^{\mathrm{gen}}_t,\tilde L^{\mathrm{acc}}_t$; set  $k_{t+1}$ via \eqref{eq:adl}.\;
}
\end{algorithm}

\subsection{Training-free, serving-compatible framework}
\label{sec:method:whole}
As summarized in Fig.~\ref{fig:diffuspec-framework}b and Alg.~\ref{alg:diffuspec}, each speculative step in DiffuSpec follows a fixed four-stage pipeline with no changes to the target model and only minimal serving-stack adjustments:
\emph{(i) Drafting} with a pretrained DLM to produce a length-$k_t$ block and per-position candidates;
\emph{(ii) CPS} on a pruned candidate lattice to select a left-to-right path aligned with AR causality;
\emph{(iii) Parallel verification} by the target $p_\theta$ (using $q_\phi^{\mathrm{L2R}}$ in the acceptance ratio) to return the accepted prefix length $L^{\mathrm{acc}}_t$ and advance the prefix;
\emph{(iv) ADL} to update the next draft length $k_{t+1}$ from the signal $L^{\mathrm{gen}}_t$ and verifier feedback $L^{\mathrm{acc}}_t$, within guardrails $[k_{\min},k_{\max}]$.
By improving the acceptance profile via CPS and right-sizing proposals via ADL, DiffuSpec increases $L^{\mathrm{acc}}_t$ per step while keeping drafting cost near the speed–quality sweet spot.
For correctness, when the verifier applies the standard speculative-decoding acceptance rule with $q_\phi^{\mathrm{L2R}}$, the classical unbiasedness analysis w.r.t.\ $p_\theta$ applies.

%% file: sec/4.experiments.tex
\section{Experiments}
\label{sec:experiments}

\textbf{Datasets.}
We follow the Spec-Bench protocol \citep{xia2024unlocking} and span six task families:
\emph{Multi-turn Conversation} (MT; \citealp{zheng2023judging}),
\emph{Machine Translation} (Trans),
\emph{Summarization} (Sum; \citealp{nallapati2016abstractive}),
\emph{Open-domain QA} (QA; \citealp{kwiatkowski2019natural}),
\emph{Mathematical Reasoning} (Math; \citealp{cobbe2021training}),
and \emph{Retrieval-Augmented Generation} (RAG; \citealp{karpukhin2020dense}). For additional details on the datasets, see Appendix~\ref{app:setup}.

\textbf{Speed metrics.}
We report (i) \emph{Mean Accepted Tokens (MAT)}, the expected length of consecutively accepted prefixes per speculative step, averaged over all steps and examples; and (ii) \emph{Speedup}, defined as end-to-end throughput relative to the AR-greedy baseline on the same target model and hardware.
All timings are wall-clock and account for DLM drafting, CPS, ADL, and parallel verification.
To ensure comparable quality (quality-locked setting), verification is performed with greedy decoding (temperature $=0$), yielding task metrics statistically indistinguishable from AR-greedy.

\textbf{Baselines.}
Our comparison covers both \emph{training-free} and \emph{training-based} speculative methods. For \emph{training-free} methods, we evaluate SPS \citep{leviathan2023fast}, Lookahead \citep{fu2024break}, PLD \citep{saxena2023prompt}, Recycling \citep{wiprachtiger2022turning}, and SAMD \citep{hu2024sam}. For \emph{training-based} systems, we report Medusa \citep{cai2024medusa}, Hydra \citep{ankner2024hydra}, and EAGLE/EAGLE2 \citep{li2024eagle,li2024eagle2}, excluding EAGLE3 \citep{li2025eagle3} due to the absence of a compatible checkpoint for our primary target. For clarity, results from the two classes are reported separately.

\textbf{Targets and drafters.}
Unless otherwise noted, the target AR model $p_\theta$ is Qwen2.5-32B~\citep{xu2025qwen2} for all training-free methods, including ours.
DiffuSpec uses Dream-7B as a training-free diffusion drafter (tokenizer aligned with Qwen2.5).
For SPS, we follow its standard AR drafter Qwen2.5-7B.
Other training-free baselines (Lookahead, PLD, Recycling, SAMD) do not employ a separate drafter.
For \emph{training-based} systems (Medusa, Hydra, EAGLE/EAGLE2), compatible Qwen2.5-32B checkpoints are unavailable; we therefore report authors’ official results on Vicuna-33B~\citep{zheng2023judging} (similar parameter scale).

\textbf{Implementation details.}
Experiments run on a single NVIDIA A100 (80GB) with 11 CPU cores and 100GB RAM, PyTorch 2.6.0.
Following \citet{kou2024cllms,wiprachtiger2022turning}, verification uses greedy decoding with batch size $=1$, KV cache enabled.
Unless stated, DiffuSpec uses a single diffusion refinement step ($S{=}1$) to isolate drafting cost.
Controller and search hyperparameters are fixed across tasks:
$k_{\min}{=}20$, $k_{\max}{=}30$,
beam size $B{=}3$,
mass threshold $\tau{=}0.8$,
per-position cap $M_{\max}{=}15$,
mixing weight $\lambda{=}0.5$,
controller increment $\delta{=}10$ tokens,
and EMA smoothing $\rho{=}0.5$.
The causal proxy is a 3-gram KenLM fitted on the training split of each dataset.

\begin{table}[t]
\centering
\resizebox{\linewidth}{!}{%
\begin{tabular}{ccccccccc}
\toprule
\multirow{2}{*}{\textbf{Model}} &
\multicolumn{6}{c}{\textbf{Speedup} ($\times$ vs.\ AR, $\uparrow$)} &
\multirow{2}{*}{\textbf{Mean (MAT / Speedup)}} \\
\cmidrule(lr){2-7}
 & \textbf{MT} & \textbf{Trans} & \textbf{Sum} & \textbf{QA} & \textbf{Math} & \textbf{RAG} & \\
\midrule
Lookahead   & 1.37$\times$ & 1.16$\times$ & 1.15$\times$ & 1.33$\times$ & 1.52$\times$ & 1.21$\times$ & 1.82 \,/\, 1.30$\times$ \\
PLD         & 1.83$\times$ & 1.29$\times$ & 2.76$\times$ & 1.87$\times$ & 1.55$\times$ & 2.37$\times$ & 2.11 \,/\, 1.93$\times$ \\
Recycling   & 2.15$\times$ & 1.85$\times$ & 2.03$\times$ & 2.06$\times$ & 2.45$\times$ & 1.83$\times$ & 3.13 \,/\, 2.07$\times$ \\
SAMD        & 1.99$\times$ & 1.54$\times$ & \textbf{3.38}$\times$ & 2.44$\times$ & 1.63$\times$ & \textbf{3.27}$\times$ & 2.18 \,/\, 2.35$\times$ \\
SPS         & 1.69$\times$ & 1.64$\times$ & 1.74$\times$ & 1.50$\times$ & 1.86$\times$ & 1.62$\times$ & 6.18 \,/\, 1.67$\times$ \\
\textbf{DiffuSpec} & \textbf{3.09}$\times$ & \textbf{3.38}$\times$ & 2.41$\times$ & \textbf{3.03}$\times$ & \textbf{4.02}$\times$ & 2.38$\times$ & \textbf{6.99 \,/\, 3.08}$\times$ \\
\midrule
Medusa      & 1.69$\times$ & 1.61$\times$ & 2.24$\times$ & 1.74$\times$ & 2.35$\times$ & 2.48$\times$ & 2.31 \,/\, 2.02$\times$ \\
Hydra       & 2.48$\times$ & 2.08$\times$ & 2.57$\times$ & 2.14$\times$ & 3.25$\times$ & 2.74$\times$ & 3.23 \,/\, 2.54$\times$ \\
EAGLE       & 2.68$\times$ & 2.21$\times$ & 2.68$\times$ & 2.24$\times$ & 3.26$\times$ & 2.50$\times$ & 3.37 \,/\, 2.76$\times$ \\
EAGLE2      & 3.45$\times$ & 2.49$\times$ & 2.94$\times$ & 2.52$\times$ & 3.70$\times$ & 2.58$\times$ & 4.02 \,/\, 2.95$\times$ \\
\bottomrule
\end{tabular}
}%
\caption{\textbf{Main results on Spec-Bench.}
Per-task columns report \emph{Speedup} only (unitless ratio vs.\ AR, $\uparrow$); the rightmost column reports the task-macro \emph{Mean (MAT / Speedup)}.
Training-free (top block) and training-based (bottom block) results use different targets.}

\label{tab:main}

\end{table}

\subsection{Effectiveness}
\label{sec:results}

\textbf{Training-free comparison.}
Tab.~\ref{tab:main} shows that, on Spec-Bench, DiffuSpec achieves the best \emph{training-free} average with Mean-MAT $6.99$ and Mean-Speedup $3.08\times$.
Compared to strong baselines, DiffuSpec improves both acceptance and wall-clock efficiency (e.g., vs.\ SPS: \,+$0.81$ MAT and \,+$1.41\times$ speedup).
At the task level, DiffuSpec attains the highest speedups on \emph{MT/Trans/QA/Math} with \(\,3.09\times/3.38\times/3.03\times/4.02\times\), indicating consistently longer accepted prefixes and faster end-to-end progress at matched quality.

\textbf{Training-based systems (context only).}
We report the results of Medusa/Hydra/EAGLE/EAGLE2 for contextual reference only, as they rely on different target models and decoding stacks.

Although not directly comparable, their metrics are on a similar scale (e.g., EAGLE2 achieves a Mean-MAT of 4.02 and a Mean-Speedup of 2.95$\times$). This suggests that diffusion-based drafting can approach training-based efficiency without extra training or serving changes.

\begin{figure}[t]
\centering
\begin{minipage}{0.48\linewidth}
  \centering
  \includegraphics[width=\linewidth]{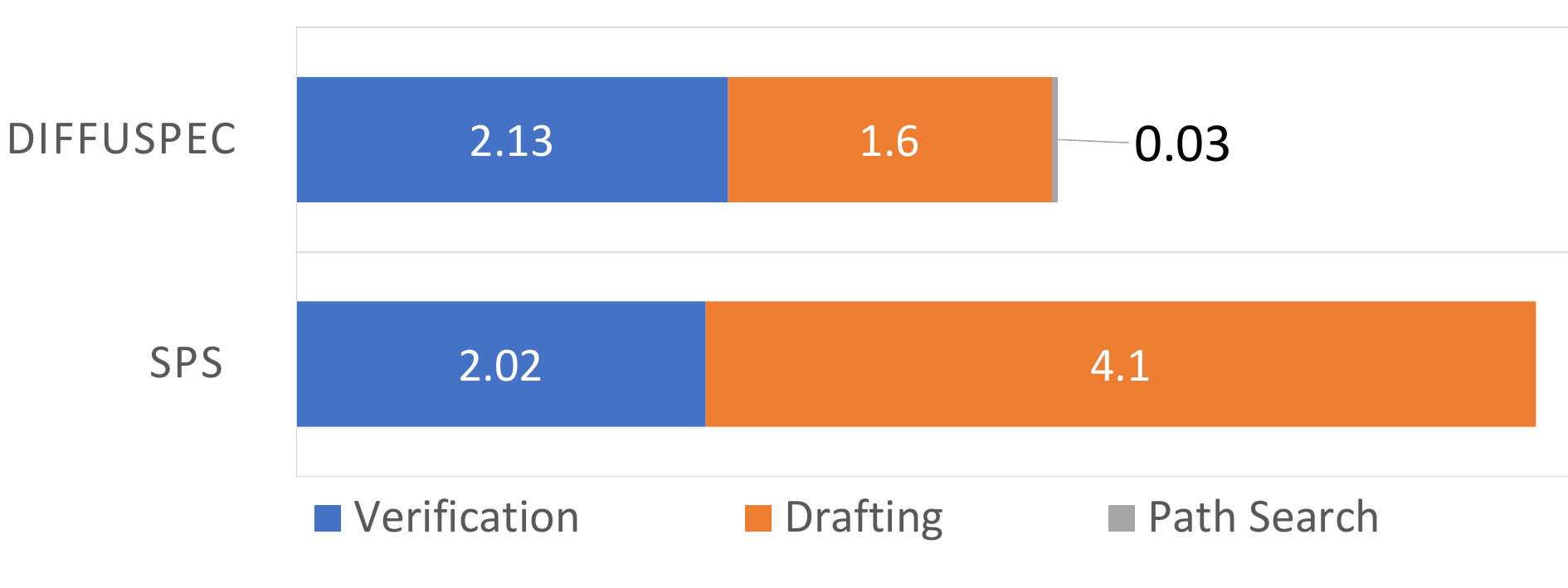} 
\captionof{figure}{\textbf{Per-step wall-clock time (s).} Mean seconds per drafter–verifier round spent in drafting, verification, and CPS (SPS vs.\ DiffuSpec).}

  \label{fig:time}
\end{minipage}\hfill
\begin{minipage}{0.48\linewidth}
  \centering
  \resizebox{\linewidth}{!}{%
  \begin{tabular}{cccc}
  \toprule
  \textbf{CPS} & \textbf{ADL} & \textbf{Mean-MAT} & \textbf{Mean-Speedup} \\
  \midrule
  \ding{51} & \ding{51} & \textbf{6.99} & \textbf{3.08$\times$} \\
  \ding{51} & \ding{55} & 6.95 & 2.98$\times$ \\
  \ding{55} & \ding{51} & 6.43 & 2.73$\times$ \\
  \ding{55} & \ding{55} & 6.05 & 2.69$\times$ \\
  \bottomrule
  \end{tabular}
  }
  \captionof{table}{\textbf{Ablation on DiffuSpec components.} \ding{51} indicates the component is enabled.
  Both ADL and CPS improve performance, with CPS contributing the larger share of gains.}
  \label{tab:ablation}
  
\end{minipage}
\vspace{-0.3cm}
\end{figure}

\textbf{Where the speedup comes from.}
Fig.~\ref{fig:time} decomposes wall-clock time into \emph{drafting}, \emph{verification}, and \emph{CPS search}.
DiffuSpec reduces drafting cost relative to SPS by using a single DLM forward pass to propose multiple tokens, while CPS adds only minor overhead (averaging $1.1\%$ across tasks).
In our setup, SPS employs a 7B AR drafter close to the target’s capacity; the resulting sequential passes dominate wall-clock and blunt the benefits of verifier parallelism—MAT remains relatively high, yet end-to-end speedup is modest.
By contrast, DiffuSpec with Dream-7B achieves substantially larger speedups at comparable or higher MAT by combining two levers: (i) higher per-step drafting throughput (non-AR DLM pass) and (ii) higher acceptance via \emph{CPS}, with \emph{ADL} right-sizing proposals.
Together, these mechanisms translate acceptance gains into tangible wall-clock acceleration.

\subsection{Ablation}

Tab.~\ref{tab:ablation} quantifies the contributions of CPS and ADL.
Enabling either module improves both \emph{Mean-MAT} and \emph{Mean-Speedup} over the plain variant, while enabling both yields the best overall performance (6.99 MAT, $3.08\times$).
Compared to the plain system (6.05 / $2.69\times$), \textit{CPS-only} raises MAT by \(+0.90\) and speedup by \(+0.29\times\), whereas \textit{ADL-only} adds \(+0.38\) MAT and \(+0.04\times\), respectively.
Thus, CPS accounts for most acceptance gains—consistent with its role in aligning diffusion proposals with AR causality—while ADL primarily translates these gains into wall-clock speedup by adaptively setting \(k_t\).
When combined, they deliver a total improvement of \(+0.39\times\) over the plain system.
Additional analysis of draft-length choices is provided in Appendix~\ref{app:fixedk}, and task-wise ablations with full results appear in Appendix~\ref{app:ablation} (Tab.~\ref{tab:app-ablation}).

\begin{figure}[t]
\centering
\begin{subfigure}{0.24\linewidth}
  \centering
  \includegraphics[width=\linewidth]{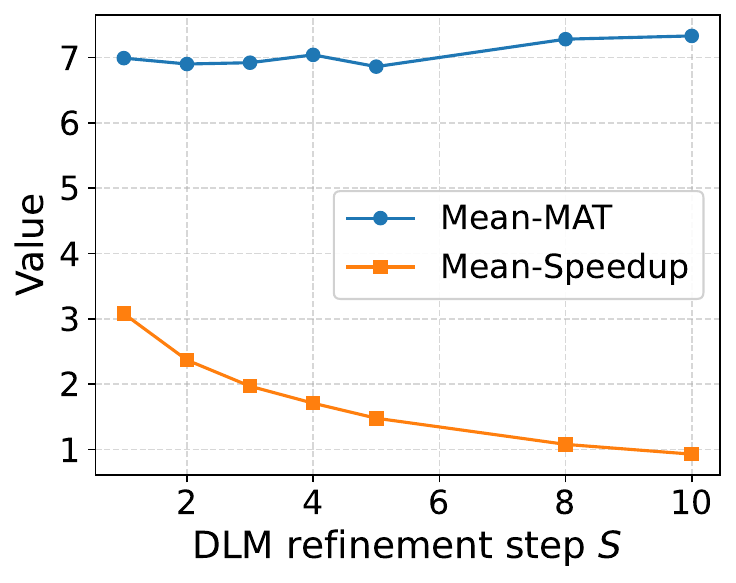}
  \caption{$S$ refinement steps}
  \label{fig:sweep_s}
\end{subfigure}\hfill
\begin{subfigure}{0.24\linewidth}
  \centering
  \includegraphics[width=\linewidth]{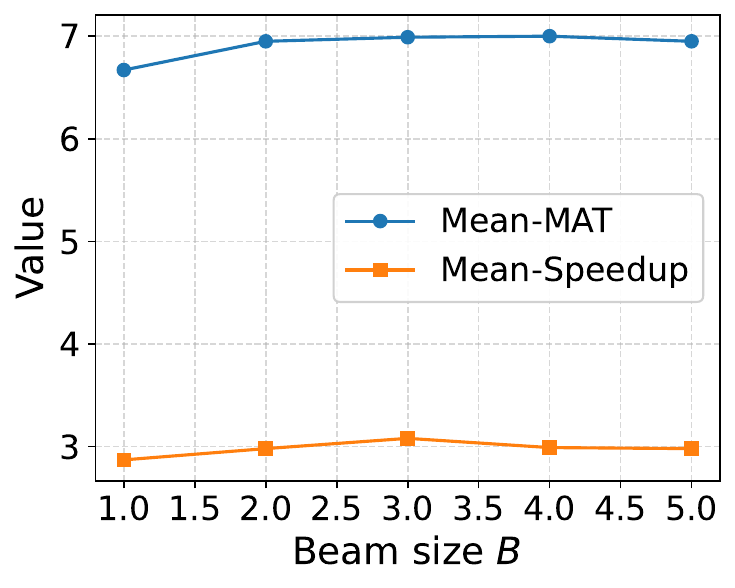}
  \caption{$B$ beam size}
  \label{fig:sweep_b}
\end{subfigure}\hfill
\begin{subfigure}{0.24\linewidth}
  \centering
  \includegraphics[width=\linewidth]{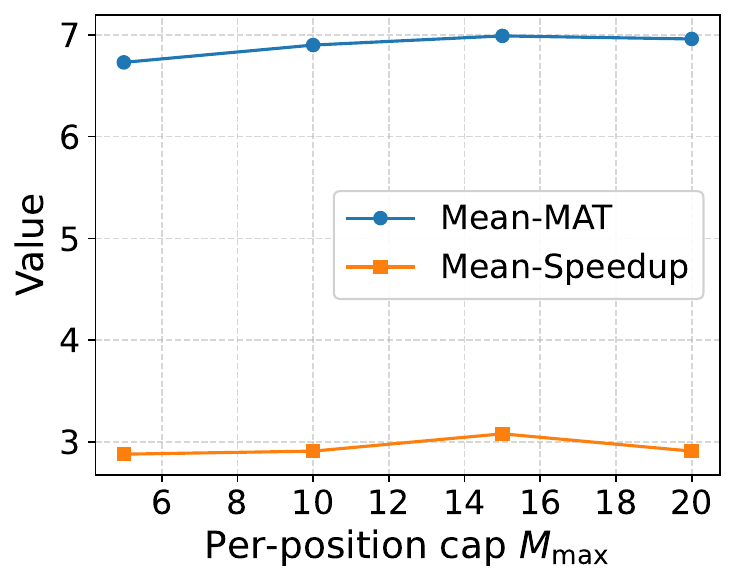}
  \caption{$M_{\max}$ per-position cap}
  \label{fig:sweep_mmax}
\end{subfigure}\hfill
\begin{subfigure}{0.24\linewidth}
  \centering
  \includegraphics[width=\linewidth]{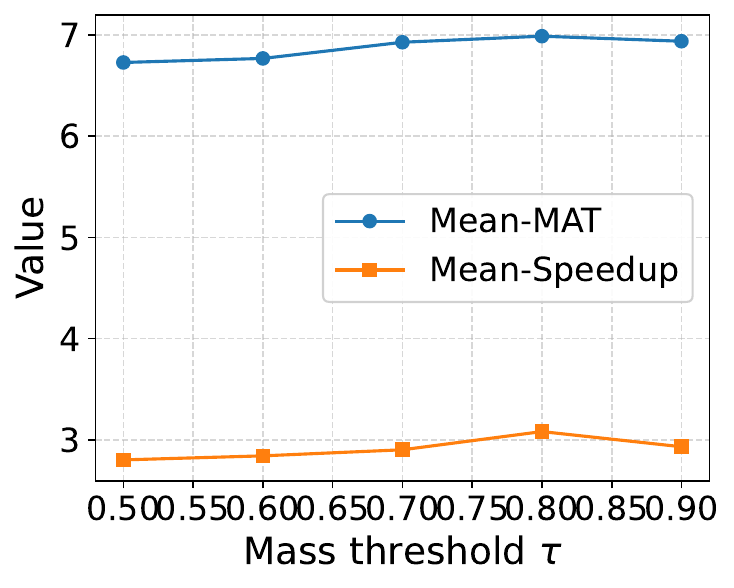}
  \caption{$\tau$ mass threshold}
  \label{fig:sweep_tau}
\end{subfigure}
\caption{\textbf{Sensitivity to decoding/search hyperparameters.} Each panel plots \emph{Mean-MAT} and \emph{Mean-Speedup} versus a single knob under the quality-locked setting.}
\label{fig:param_sweep}
\end{figure}

\subsection{Hyperparameter sensitivity}
\label{sec:hyperparam}
Across decoding/search knobs (Fig.~\ref{fig:sweep_s}–\ref{fig:sweep_tau}), we observe consistent speed–quality trade-offs under the quality lock. Increasing the number of DLM refinement steps $S$ improves proposal quality and acceptance (Mean-MAT \(6.99\rightarrow7.33\) from $S{=}1$ to $10$; Fig.~\ref{fig:sweep_s}), but substantially reduces throughput (Mean-Speedup \(3.08\times\rightarrow0.93\times\)), so we fix $S{=}1$. Enlarging the CPS beam $B$ improves causal paths and modestly raises Mean-MAT, peaking around $B{=}3{\sim}4$ (Fig.~\ref{fig:sweep_b}); however, overhead causes speedup to plateau or regress beyond $B{=}3$, so we set $B{=}3$. Increasing the per-position cap $M_{\max}$ relaxes pruning and helps until $M_{\max}{\approx}15$ (Fig.~\ref{fig:sweep_mmax}); further branching yields negligible gains and slightly hurts speed, motivating our choice of $M_{\max}{=}15$. Raising the mass threshold $\tau$ retains more local probability and improves acceptance/speed up to $\tau{\approx}0.8$ (Fig.~\ref{fig:sweep_tau}); higher values add compute with little benefit, so we use $\tau{=}0.8$. Overall, CPS-related knobs ($B, M_{\max}, \tau$) are robust over a broad range, while multi-step refinement $S$ trades acceptance for latency. Orthogonally, ADL controls proposal size, helping convert CPS-driven acceptance gains into wall-clock acceleration.

%% file: sec/6.conclusion.tex
\section{Conclusion and Future Work}
\label{sec:conclusion}
We introduced \textbf{DiffuSpec}, a training-free drop-in framework for speculative decoding that employs a diffusion language model (DLM) as the drafter. To reconcile diffusion-based drafting with AR verification, we proposed \emph{causal-consistency path search} (CPS) and an \emph{adaptive draft-length} (ADL) controller. Across six task families, DiffuSpec produces high-quality multi-token drafts, delivering the strongest speedups among training-free baselines and approaching training-based systems under quality-locked settings. Ablations indicate that both CPS and ADL improve acceptance and throughput: CPS yields the larger gains by aligning proposals with AR causality, whereas ADL stabilizes proposal size to avoid over-/under-drafting. DiffuSpec requires no additional neural training and integrates with existing targets with minimal serving-stack changes.

For future work, we highlight three directions: (i) \textit{system-level acceleration} for DLM drafting (e.g., KV-cache–style reuse and fused kernels); (ii) \textit{stronger proposal selection} via improved causal proxies or verifier-aware scoring to further increase acceptance; and (iii) \textit{richer adaptive control} that jointly tunes draft length and search/pruning breadth online. We hope DiffuSpec provides a practical blueprint for bridging diffusion-based generation with fast verifier-aligned decoding.

%% file: sec/appendix.tex
\newpage
\appendix

\section{Dataset and Implementation Details}
\label{app:setup}

\paragraph{Datasets.}
We follow \textsc{Spec-Bench} \citep{xia2024unlocking} across six task families, using the official splits and preprocessing; prompts match \S\ref{sec:experiments}. For \emph{Multi-turn Conversation} (MT) we use \textbf{MT-Bench} with pairwise judging \citep{zheng2023judging}. \emph{Machine Translation} (Trans) follows Spec-Bench’s public WMT-style news configuration. \emph{Summarization} (Sum) is \textbf{CNN/DailyMail} \citep{nallapati2016abstractive}. \emph{Open-domain QA} (QA) is \textbf{Natural Questions} \citep{kwiatkowski2019natural}. \emph{Mathematical Reasoning} (Math) uses \textbf{GSM8K} \citep{cobbe2021training}. \emph{Retrieval-Augmented Generation} (RAG) follows the \textbf{DPR} pipeline over Wikipedia \citep{karpukhin2020dense}.

\begin{table}[h]
\centering
\resizebox{0.85\linewidth}{!}{%
\begin{tabular}{lll}
\toprule
\textbf{Task} & \textbf{Dataset(s)} & \textbf{Metric(s)} \\
\midrule
MT    & MT-Bench \citep{zheng2023judging}                 & Win rate (pairwise) \\
Trans & Spec-Bench translation (WMT-style)                & BLEU \\
Sum   & CNN/DailyMail \citep{nallapati2016abstractive}    & ROUGE-L \\
QA    & Natural Questions \citep{kwiatkowski2019natural}  & EM / F1 \\
Math  & GSM8K \citep{cobbe2021training}                   & Accuracy \\
RAG   & DPR over Wikipedia \citep{karpukhin2020dense}     & Accuracy \\
\bottomrule
\end{tabular}
}
\caption{Spec-Bench datasets and evaluation metrics used in our experiments.}
\label{tab:app-datasets}
\vspace{-0.3cm}
\end{table}

\paragraph{Implementation details.}
We build our evaluation harness on top of \textsc{Spec-Bench} \citep{xia2024unlocking}, reusing its official data loaders, prompt templates and stop criteria. All systems share the same hardware/software stack as \S\ref{sec:experiments} (single NVIDIA A100~80GB, 11 CPU cores, 100GB RAM, PyTorch~2.6.0). Verification uses greedy decoding (temperature $=0$) with batch size $=1$ and KV cache enabled; we report tokens/s averaged over the full evaluation set, excluding model-loading and first-batch warmup. Wall-clock timing includes tokenization, drafter forward(s), path search, verifier forward, and residual sampling.

Unless otherwise stated, DiffuSpec adopts a single diffusion refinement step ($S{=}1$). Controller and search hyperparameters are fixed across tasks:
$k_{\min}{=}20$, $k_{\max}{=}30$, beam size $B{=}3$, mass threshold $\tau{=}0.8$, per-position cap $M_{\max}{=}15$, mixing weight $\lambda{=}0.5$, controller increment $\delta{=}10$, and EMA smoothing $\rho{=}0.5$. The causal proxy is a 3-gram KenLM  fitted \emph{only} on the training split of each dataset (no test leakage). Speedup is defined as a unitless ratio: throughput(method) / throughput(AR-greedy) under identical runtime settings; MAT follows Sec.~\ref{sec:prelim}. CUDA events are synchronized at measurement points to ensure consistent timing.

\section{Full Ablation Results per Task}
\label{app:ablation}

Table~\ref{tab:app-ablation} expands Table~\ref{tab:ablation} by reporting 
task-wise MAT and Speedup under different combinations of 
causal-consistency path search (CPS) and adaptive draft-length (ADL).

\begin{table}[h]
\centering
\resizebox{\linewidth}{!}{%
\begin{tabular}{cccccccccccccc}
\toprule
\multirow{2}{*}{\textbf{CPS}} & \multirow{2}{*}{\textbf{ADL}} 
& \multicolumn{2}{c}{\textbf{MT}} & \multicolumn{2}{c}{\textbf{Trans}} 
& \multicolumn{2}{c}{\textbf{Sum}} & \multicolumn{2}{c}{\textbf{QA}} 
& \multicolumn{2}{c}{\textbf{Math}} & \multicolumn{2}{c}{\textbf{RAG}} \\
\cmidrule(lr){3-14}
& & MAT & Spd & MAT & Spd & MAT & Spd & MAT & Spd & MAT & Spd & MAT & Spd \\
\midrule
\ding{51} & \ding{51} & \textbf{7.02} & \textbf{3.12}$\times$ & \textbf{7.35} & \textbf{3.40}$\times$ & \textbf{6.25} & 2.45$\times$ & \textbf{7.49} & \textbf{3.05}$\times$ & \textbf{7.61} & \textbf{4.05}$\times$ & \textbf{8.04} & 2.40$\times$ \\
\ding{51} & \ding{55} & 6.98 & 3.01$\times$ & 7.29 & 3.28$\times$ & 6.18 & 2.37$\times$ & 7.40 & 2.92$\times$ & 7.55 & 3.88$\times$ & 7.96 & 2.34$\times$ \\
\ding{55} & \ding{51} & 6.41 & 2.70$\times$ & 6.72 & 2.79$\times$ & 5.88 & 2.11$\times$ & 6.92 & 2.55$\times$ & 7.00 & 3.11$\times$ & 7.25 & 2.14$\times$ \\
\ding{55} & \ding{55} & 6.03 & 2.65$\times$ & 6.48 & 2.61$\times$ & 5.72 & 1.96$\times$ & 6.75 & 2.41$\times$ & 6.82 & 2.95$\times$ & 7.08 & 2.02$\times$ \\
\bottomrule
\end{tabular}
}
\caption{\textbf{Task-wise ablation of DiffuSpec components.}
CPS = causal-consistency path search; ADL = adaptive draft-length.
Both components improve MAT and speedup (Spd) across tasks; \textbf{Spd denotes Speedup} ($\times$ vs.\ AR, $\uparrow$).}
\label{tab:app-ablation}
\vspace{-0.3cm}
\end{table}

The task-wise breakdown confirms the complementary roles of CPS and ADL. 
CPS consistently yields larger gains, especially on QA and Math where 
alignment with AR verification is critical. ADL offers steady 
improvements by preventing over/under-drafting, with a visible effect 
on Summarization. Combining both mechanisms produces the best overall 
results, robust across all  tasks.

\section{Fixed Draft Length Study}
\label{app:fixedk}

We evaluate fixed proposal lengths $k\!\in\!\{10,20,30,50,100\}$ as well as the adaptive controller (ADL). Table~\ref{tab:app-fixedk}  shows the trade-off: longer drafts increase acceptance length but reduce throughput due to higher rejection rates and drafting overhead.

\begin{table}[h]
\centering
\resizebox{0.8\linewidth}{!}{%
\begin{tabular}{lcccccc}
\toprule
\textbf{Policy} & \textbf{$k{=}10$} & \textbf{$k{=}20$} & \textbf{$k{=}30$} & \textbf{$k{=}50$} & \textbf{$k{=}100$} & \textbf{ADL} \\
\midrule
Mean-MAT     & 5.53 & 6.56 & 6.49 & 6.51 & 6.69 & \textbf{6.99} \\
Mean-Speedup & 2.74$\times$ & 2.98$\times$ & 2.98$\times$ & 2.91$\times$ & 2.78$\times$ & \textbf{3.08$\times$} \\
\bottomrule
\end{tabular}
}
\caption{\textbf{Fixed-$k$ vs.\ adaptive proposal length (quality-locked).} Means are computed across all tasks.
ADL achieves the best speedup while also reaching the highest MAT, indicating a better speed–acceptance trade-off than fixed-$k$ policies.}
\label{tab:app-fixedk}
\end{table}

As $k$ increases from 10 to 100, Mean-MAT generally rises (peaking at $k{=}100$ with 6.69), but Mean-Speedup peaks earlier at $k{=}20/30$ (both 2.98$\times$) and then declines due to higher drafting and rejection costs. The adaptive controller (ADL) balances this trade-off online, attaining both the highest Mean-MAT (6.99) and the strongest Mean-Speedup (3.08$\times$). This confirms the benefit of dynamic proposal sizing over fixed-$k$ policies.

\section{Output Visualizations}
\label{app:viz}

We provide qualitative $k$-sweeps showing how draft length shapes proposal style: short drafts tend to be terse; moderate drafts begin to exhibit step-by-step reasoning; very long drafts may drift or repeat. (All runs use the same prompt; only $k$ varies. Visualization samples are raw drafts before CPS/verification and are not correctness-checked.)

\begin{tcolorbox}[title={Algebra: \emph{Solve for $x$: $3x+5=20$}}, colback=white,colframe=black!15]
\noindent \textbf{$k{=}6$:}\; \texttt{Subtract 5 and divide by 3.}\\
\textbf{$k{=}60$:}\;\texttt{To solve for $x$, subtract 5 from both sides and then divide by 3 to isolate $x$.}\\
\textbf{$k{=}200$:}\; \texttt{To solve for $x$, we first isolate the variable by removing constants and normalizing the coefficient of $x$ \dots}
\end{tcolorbox}

\vspace{0.4em}
\begin{tcolorbox}[title={Word problem: \emph{A train travels at 60 km/h for 2.5 hours. How far?}}, colback=white,colframe=black!15]
\noindent \textbf{$k{=}6$:}\; \texttt{150 km.}\\
\textbf{$k{=}20$:}\; \texttt{Distance = speed $\times$ time = $60\times2.5=150$ km.}\\
\textbf{$k{=}60$:}\;\texttt{To find the distance, multiply the constant speed by the elapsed time; units remain in km.}
\end{tcolorbox}

As $k$ increases, drafts shift from terse answers to step-by-step reasoning (often with emerging chain-of-thought), which \emph{initially} raises the verifier’s accepted length: MAT grows for small-to-moderate $k$.
Beyond a task-dependent sweet spot, however, we observe a clear \emph{plateau}: very long drafts tend not to yield longer accepted prefixes—diffusion proposals begin to drift, repeat, or include partial phrases, so the AR verifier rejects earlier.
Consequently, end-to-end speedup drops due to extra drafting and residual resampling, even though the draft itself is longer.
This motivates \emph{adaptive} proposal sizing (ADL) to stay near the knee of the MAT/speed trade-off, and \emph{causal-consistency} path search (CPS) to keep proposals informative yet easy for the verifier to accept.